\documentclass[12pt]{article}
\usepackage{amsmath,amssymb,bm,graphicx,algorithm,algorithmic}
\usepackage[top=25truemm,bottom=25truemm,left=25truemm,right=25truemm]{geometry}
\allowdisplaybreaks

\title{Automatic Hyperparameter Tuning in Sparse Matrix Factorization}

\author{Ryota Kawasumi\thanks{Department of Mathematics, Graduate School of Science and Engineering, Chuo University, e-mail: rykawasumi@gmail.com} 
\and Koujin Takeda\thanks{Department of Mechanical Systems Engineering, Graduate School of Science and Engineering, Ibaraki University, e-mail: koujin.takeda.kt@vc.ibaraki.ac.jp}}

\date{4th January, 2023}

\begin{document}
\maketitle

\begin{abstract}
We study the problem of hyperparameter tuning in sparse matrix factorization under Bayesian framework.
In the prior work, an analytical solution of sparse matrix factorization with Laplace prior 
was obtained by variational Bayes method under several approximations. 
Based on this solution, we propose a novel numerical method of 
hyperparameter tuning by evaluating the zero point of normalization factor in sparse matrix prior.
We also verify that our method shows excellent performance for ground-truth sparse matrix reconstruction
by comparing it with the widely-used algorithm of sparse principal component analysis.
\end{abstract}
\vspace{3mm}


%%%%%%%%%%%%%%%%%%%%%
\section{Introduction}
\label{sec1}

Among machine learning problems, matrix factorization (MF) is significant  
because MF appears in many applications such as recommendation system, signal processing, etc.
We restrict ourselves to sparse MF problem in this article, where either factorized matrix must be sparse.
This is originally discussed as sparse coding in neuroscience \cite{OF1,OF2}, 
and recognized as a significant problem in neuronal information processing in the brain. 
It also appears in sparse modeling in information science such as 
dictionary learning \cite{EAH,AEB} or sparse principal component analysis 
(sparse PCA) \cite{ZHT1,AGJL}.

Many attempts have been made so far for understanding theoretical aspects of MF,
and analytical tools for random systems in statistical physics are found to be useful, 
e.g. Markov chain Monte Carlo method \cite{SM}, replica analysis \cite{SK, KMZ, KKMSZ, SSZ, LKZ2}, 
and message passing \cite{KMZ, KKMSZ, SSZ, LKZ2, MT, LKZ1}, where some works are not limited to sparse matrix case.
The authors of this article also analyzed sparse MF under Laplace prior (or $\ell_1$ regularizer)
by variational Bayes (VB) method for Kullback-Leibler (KL) divergence, 
or equivalently variational calculus for free energy from the viewpoint of statistical physics \cite{KT}. 
They obtained the analytical solution of sparse MF under several approximations, which serves as sparse MF algorithm. 
They also experimentally found that ground-truth sparse MF solution can be reconstructed
by this algorithm under appropriate hyperparameter value in Laplace prior.
However, hyperparameter tuning is generally difficult in machine learning. 
Several methods based on risk estimate, information criterion, or cross validation
have been proposed in well-known Lasso \cite{ZHT2,DKFPC,VDFPD,WZLMM}, some of which are based on replica analysis or message passing \cite{BEM,OK,MMB}. 
On the other hand, such method has not been sufficiently discussed for sparse MF, especially under $\ell_1$ regularizer.

Here we propose a novel numerical approach for
hyperparameter tuning in Laplace prior in sparse MF solution \cite{KT}.
The essential idea is the zero point of normalization factor.
In sparse MF solution, the contribution from Laplace prior
is expressed by the correction terms to MF solution under uniform prior.
All these terms include normalization factor of probability distribution 
for sparse matrix in their denominators. 
Hence, their contribution becomes dominant
in the vicinity of the zero point of normalization factor.
Based on this idea, we construct an MF algorithm 
including evaluation of the zero point for hyperparameter tuning.
However, sparse MF solution was obtained under several approximations, and 
our numerical approach does not necessarily lead to sparse MF solution.
Nevertheless, our experiment shows that ground-truth sparse MF solution
can be reconstructed with high accuracy.
We also compare the performance of our algorithm with widely-used sparse PCA algorithm \cite{MBPS}.
Consequently, in the case of sparser ground-truth matrix, we find that our method shows better performance 
than sparse PCA algorithm under hyperparameter tuning by hand.

%%%%%%%%%%%%%%%%%%%%%
\section{Outline of VB analysis for sparse MF}
\label{sec2}

VB analysis for sparse MF solution is outlined here.
In MF, observed matrix ${\bm V} \in {\mathbb R}^{L \times M}$ 
is represented by ${\bm V}
=
{\bm A}{\bm B} + {\bm E}$, where
${\bm A} \in {\mathbb R}^{L \times H} $,
${\bm B} \in {\mathbb R}^{H \times M} $, and noise matrix
${\bm E} \in {\mathbb R}^{L \times M}$.
Here matrix is denoted by boldface letter.
The task is to find factorized matrices ${\bm A}, {\bm B}$ from ${\bm V}$, and
Bayesian approach is taken for the purpose.
We assume multivariate Gaussian prior (or $\ell_2$ regularizer) for dense matrix ${\bm A}$ and Laplace prior 
(or $\ell_1$ regularizer) for sparse matrix ${\bm B}$,
\begin{eqnarray}
P({\bm A}) 
&\propto&
\prod_{l,h} \exp\Bigr( - \frac{ a_{lh}^2  }{2({C}_{\bm A})_{hh} } \Bigr),
\label{P1}
\\
P({\bm B})
 &\propto&
\prod_{h,m}
\exp\Bigr( - \frac{ |b_{hm}|  }{ k } \Bigr),
\label{P2}
\end{eqnarray}
where $\bm C_A$ is covariance matrix for matrix $\bm A$, and
$k$ is hyperparameter in Laplace prior.
The element in noise matrix $\bm E$ is
drawn from Gaussian distribution ${\cal N} (0, \sigma^2)$.
From Gaussianity of noise, the likelihood for sparse MF model is given by
\begin{equation}
P({\bm V}|{\bm A}, {\bm B})
\propto
\prod_{l,m}
\exp\Bigr\{ - \frac{1}{2\sigma^2} \Bigr( v_{lm} - \sum_h a_{lh}b_{hm} \Bigr)^2   \Bigr\}.
\label{P3}
\end{equation}

%%%%%%%%%%%%%%%%%%%%%

With these priors and the likelihood, the matrices $\bm A, \bm B$ can be estimated
from Bayes formula by maximization of posterior, which is however computationally infeasible.
In reference \cite{KT}, VB analysis is used to tackle this problem. 
In VB analysis, the minimum of KL divergence between trial functions and 
true posterior for the matrices $\bm A, \bm B$ is evaluated. However, unlike the case of Gaussian prior for both matrices \cite{IR,NS},
the analytical solution of KL minimum cannot be obtained without approximations. Hence, several approximations are used: 
mean field approximation, $1/k$ expansion of $P(\bm B)$ up to the first order, 
and neglect of covariance. 

By KL divergence minimization with VB analysis, 
the expressions of means $\bar{a}_{lh}, \bar{b}_{hm}$ 
and variances $({\Sigma}_{{\bm A} l} )_{hh}, (\Sigma_{{\bm B}m})_{hh}$
for trial functions of $\bm A, \bm B$ are obtained as below.
More precisely, the variables $\bar{a}_{lh}, ({\Sigma}_{{\bm A} l} )_{hh}$ represent the mean and the variance 
of $lh$-element in $\bm A$, respectively.
Similarly, the variables $\bar{b}_{hm}, (\Sigma_{{\bm B}m})_{hh}$ denote
the mean and the variance of $hm$-element in $\bm B$, respectively. 
The sums $\sum_l ({\Sigma}_{{\bm A} l} )_{hh}, \sum_m (\Sigma_{{\bm B}m})_{hh}$
are the diagonal $hh$-element of covariance matrix $\bm A^{T} \bm A$ and $\bm B \bm B^{T}$, respectively.
See reference \cite{KT} for the details of the analysis.
\begin{eqnarray}
\bar{a}_{lh}
&=&
\sum_{m,h'}
(\hat{\Sigma}_{{\bm A}l}^{-1})_{hh'} v_{lm} \bar{b}_{h'm},
\label{A1} \\
({\Sigma}_{{\bm A} l} )_{hh}
&=&
\sigma^2 (\hat{\Sigma}_{{\bm A}l}^{-1})_{hh},
\label{A2} 
\end{eqnarray}
\begin{eqnarray}
\bar{b}_{hm}
&=&
\sum_{h',l} (\hat{\Sigma}_{{\bm B} m}^{-1})_{hh'} v_{lm} \bar{a}_{lh'}  -
\sum_{h'}
\frac{ \sigma^2(\hat{\Sigma}_{{\bm B}m}^{-1})_{h' h}  }{k Z_B}
{\rm erf}\bigr(\omega_{h' m}\bigr), \label{B1} \\
(\Sigma_{{\bm B}m})_{hh}
&=&
\sigma^2 (\hat{\Sigma}_{{\bm B}m}^{-1})_{hh} -\sum_{h' }
\sqrt{\frac{2}{\pi \sigma^2 (\hat{\Sigma}_{{\bm B}m}^{-1})_{h'h'}}}
\frac{ \{\sigma^2 (\hat{\Sigma}_{{\bm B}m}^{-1})_{h' h} \}^2  }{ k Z_B  }
e^{-\omega_{h' m}^2 }
\nonumber \\
&& \hspace{2cm} - 
\Bigr\{
\sum_{h'}
\frac{ \sigma^2(\hat{\Sigma}_{{\bm B}m}^{-1})_{h' h} }{k Z_B }
{\rm erf}\bigr(\omega_{h' m}\bigr)
\Bigr\}^2,
\label{B2}
\end{eqnarray}
where ${\rm erf}(x) = (2/\sqrt{\pi}) \int_x^{\infty} \exp(-t^2) dt$
and the superscript $-1$ means inverse matrix.
The matrices $\hat{\bm \Sigma}_{{\bm A} l}, \hat{\bm \Sigma}_{{\bm B} m}
 \in {\mathbb R}^{H^2}$ are obtained as
\begin{eqnarray}
(\hat{ \Sigma}_{{\bm A}l})_{hh'}
&=&
\frac{\displaystyle \sigma^2}{\displaystyle ({ C}_{{\bm A}})_{hh}} \delta_{h,h'} 
+ \sum_m \Bigl(({ \Sigma}_{{\bm B} m})_{hh} \delta_{h,h'} +
\bar{b}_{hm}\bar{b}_{h'm}
\Bigr), \label{hA2}\\
(\hat{\Sigma}_{{\bm B} m})_{hh'}
&=&
\sum_l \Bigl(({ \Sigma}_{{\bm A }l})_{hh}\delta_{h,h'}+\bar{a}_{lh}\bar{a}_{lh'}\Bigr).
\label{hB2}
\end{eqnarray}
The factors $\omega_{hm}$ and $Z_B$ are defined by
\begin{eqnarray}
\omega_{hm} 
&=&
\frac{\sum_{h',l} (\hat{\Sigma}_{{\bm B}m}^{-1})_{hh'} 
v_{lm} \bar{a}_{lh'} }{\sqrt{2 \sigma^2 (\hat{\Sigma}_{{\bm B}m}^{-1})_{hh} }},  
\label{omega}
\end{eqnarray}
\begin{eqnarray}
Z_B &=&
1 - \frac{1}{k} \sum_{m, h} \!
\left\{ \!
\sqrt{ \frac{ 2 \sigma^2 (\hat{\Sigma}_{{\bm B} m}^{-1})_{hh} }{ \pi}  } 
e^{-\omega_{h m}^2 } +
\Bigr(
\sum_{h',l} (\hat{\Sigma}_{{\bm B} m}^{-1})_{h h'} v_{lm} \bar{a}_{lh'}
\Bigr)  
 {\rm erf} \bigr(\omega_{h m}\bigr) \! \right\}. \label{ZB} \nonumber \\
\end{eqnarray}
The factor $Z_B$ serves as the normalization factor for the probability $P(\bm B)$.
Note that $Z_B$ may become negative or ill-defined for very small $k$ due to $1/k$ expansion.

By evaluating the means and the matrices in equations (\ref{A1})-(\ref{B2}) 
in conjunction with equations (\ref{hA2})-(\ref{ZB}) numerically,
we can obtain the inferred factorized matrices $\bm A, \bm B$ from $\bar{a}_{lh}, \bar{b}_{hm}$. 
As suggested in reference \cite{KT}, the performance of sparse matrix reconstruction 
depends on the value of $k$, and appropriate value of $k$ will exist.

%%%%%%%%%%%%%%%%%%%%%
\section{The idea for hyperparameter tuning}
\label{sec3}

In our approach for hyperparameter tuning, we focus on the correction terms in sparse MF solution equations (\ref{A1})-(\ref{B2}).
The key is the zero point of normalization factor $Z_B$.
In numerical experiment in reference \cite{KT}, the zero point of $Z_B$ is close to optimal $k$ for sparse matrix reconstruction.
The reason will be as follows.
In equations (\ref{B1}) and (\ref{B2}), all $k$-dependent terms originated from Laplace prior have $1/k Z_B$ factor.
Hence, these terms will become dominant in the vicinity of the zero point of $Z_B$, whereas
they disappear in the limit of $k \to \infty$ or uniform prior case for $\bm B$.
From equation (\ref{ZB}), the value of $k$ at the zero point of $Z_B$ is given by
\begin{eqnarray}
k \!\! &=& \!\!\!
\sum_{m,h}
\biggr\{
\sqrt{ \frac{ 2\sigma^2 (\widehat{ \Sigma}_{{\bm B}m}^{-1})_{hh} }{\pi}  } 
e^{-\omega_{hm}^2}  +
\Bigr(
\sum_{h',l} v_{lm}(\widehat{ \Sigma}_{{\bm B}m}^{-1})_{hh'}  \bar{a}_{lh'}
\Bigr) \
 {\rm erf} \bigr(\omega_{hm}\bigr)
\biggr\}. \label{ZBzero}
\end{eqnarray}
Note that the variables $\bar{a}_{lh'}$,$(\widehat{ \Sigma}_{{\bm B}m}^{-1})_{hh'}$, $\omega_{hm}$
on r.h.s. depend on $k$. Therefore, 
we need to solve this equation numerically to evaluate the zero point of $Z_B$.
 
The simplest idea for the zero point is to iteratively update $k$ by computing r.h.s of equation (\ref{ZBzero}). 
However, simple iteration for $k$ leads to instability. We should introduce partial update of $k$ for convergence,
\begin{eqnarray}
k^{(t+1)} &=& (1-\epsilon) k^{(t)} \nonumber \\
&& \hspace{-5mm} + \epsilon \sum_{m,h}
\biggr\{
\sqrt{ \frac{ 2\sigma^2 (\widehat{ \Sigma}_{{\bm B}m}^{-1})_{hh} }{\pi}  } 
e^{-\omega_{hm}^2}  + 
\Bigr(
\sum_{h',l} v_{lm}(\widehat{ \Sigma}_{{\bm B}m}^{-1})_{hh'}  \bar{a}_{lh'}
\Bigr) \
 {\rm erf} \bigr(\omega_{hm}\bigr)
\biggr\}, \label{updatek} \nonumber \\
\end{eqnarray}
where $\epsilon$ is an arbitrary small constant for partial update, and the superscript $(t)$ denotes iteration step.

By considering equations (\ref{A1})-(\ref{ZB}) and (\ref{updatek}) as an iterative algorithm, we can construct a
sparse MF algorithm as in algorithm \ref{alg1}. 
For the termination condition of the algorithm, we use the threshold $Z_{B {\rm thres}}$ for normalization factor $Z_B$. 

We should remember that sparse MF solution equations (\ref{A1})-(\ref{B2}) is obtained by the first order approximation of $1/k$,
which may lose the advantage of $\ell_1$ regularizer for sparsity.
Therefore, there is no guarantee that our approach will lead to sparse MF solution. 
We should verify the validity of our approach carefully.

\begin{algorithm}                       
\caption{iterative MF algorithm with hyperparameter tuning for sparse prior}         
\label{alg1}                          
\begin{algorithmic}
\STATE{set $t=0$}                  
\STATE{set initial $k^{(0)}$ and $Z_B^{(0)}$ very large}
\STATE{initialize $\bar{a}_{lh}^{(0)}, \bar{b}_{hm}^{(0)}$ randomly $\forall l,h,m$
}
\STATE{initialize $\Sigma_{{\bm B}m}^{(0)}$ as identity matrix $\forall m$}
\WHILE{ $Z_B^{(t)} > Z_{B {\rm thres}}$}
\STATE{$(\hat{\Sigma}_{{\bm A} l} )_{hh'}^{(t+1)} =  \frac{\displaystyle \sigma^2}{\displaystyle ({ C}_{{\bm A}})_{hh}} \delta_{h,h'} + \sum_m \Bigl(({ \Sigma}_{{\bm B} m})_{hh}^{(t)} \delta_{h,h'} +
\bar{b}_{hm}^{(t)} \bar{b}_{h'm}^{(t)} \Bigr), \ \forall h,h' $}
\STATE{$({\Sigma}_{{\bm A} l} )_{hh}^{(t+1)} = \sigma^2 (\hat{\Sigma}_{{\bm A}l}^{-1})_{hh}^{(t+1)}, \ \forall h$}
\STATE{$\bar{a}_{lh}^{(t+1)} = \sum_{m,h'} (\hat{\Sigma}_{{\bm A}l}^{-1})_{hh'}^{(t+1)} v_{lm} \bar{b}_{h'm}^{(t)}, \forall l,h$}
\STATE{$(\hat{\Sigma}_{{\bm B} m})_{hh'}^{(t+1)} \!\!=\!\! \sum_l \Bigl(({ \Sigma}_{{\bm A }l})_{hh}^{(t+1)} \delta_{h,h'}\!\! +\!\! \bar{a}_{lh}^{(t+1)} \bar{a}_{lh'}^{(t+1)} \Bigr),\forall h,h'$}

\STATE{$\omega_{hm}^{(t+1)} 
=
\frac{\sum_{h',l} (\hat{\Sigma}_{{\bm B}m}^{-1})_{hh'}^{(t+1)} 
v_{lm} \bar{a}_{lh'}^{(t+1)} }{\sqrt{2 \sigma^2 (\hat{\Sigma}_{{\bm B}m}^{-1})_{hh}^{(t+1)} }},\ \forall h,m$ }

\STATE{$k^{(t+1)} = (1-\epsilon) k^{(t)}+ \epsilon \left\{
\sum_{m,h} \Big(
\sqrt{ \frac{ 2\sigma^2 (\widehat{ \Sigma}_{{\bm B}m}^{-1})_{hh}^{(t+1)} }{\pi}  } 
e^{-(\omega_{hm}^{(t+1)})^2} \Big)  \right.$}
\STATE{$\hspace{5cm} \left. +
\big( \sum_{h',l} (\widehat{ \Sigma}_{{\bm B}m}^{-1})_{hh'}^{(t+1)} v_{lm} \bar{a}_{lh'}^{(t+1)}
\big) \ {\rm erf} \bigr( \omega_{hm}^{(t+1)} \bigr) \right\} $ }
\STATE{$ Z_B^{(t+1)} = 1 - \frac{1}{k^{(t+1)}} \sum_{m, h}
\left\{
\sqrt{ \frac{ 2 \sigma^2 (\hat{\Sigma}_{{\bm B} m}^{-1})_{hh}^{(t+1)} }{ \pi}  } 
e^{- (\omega_{h m}^{(t+1)})^2 } \right.$}
\STATE{\hspace{5cm} $\left. + \Bigr(
\sum_{h',l} (\hat{\Sigma}_{{\bm B} m}^{-1})_{h h'}^{(t+1)} v_{lm} \bar{a}_{lh'}^{(t+1)}
\Bigr) \ 
 {\rm erf} \bigr(\omega_{h m}^{(t+1)} \bigr) \right\}$ }
\STATE{$(\Sigma_{{\bm B}m})_{hh}^{(t+1)} = \sigma^2 (\hat{\Sigma}_{{\bm B}m}^{-1})_{hh}^{(t+1)}
 - \sum_{h'} \sqrt{\frac{2}{\pi \sigma^2 (\hat{\Sigma}_{{\bm B}m}^{-1})_{h'h'}^{(t+1)}}}
\frac{ \{\sigma^2 (\hat{\Sigma}_{{\bm B}m}^{-1})_{h' h}^{(t+1)} \}^2  }{ k^{(t+1)} Z_B^{(t+1)}  }
e^{- (\omega_{h' m}^{(t+1)})^2 }$}
\STATE{\hspace{5cm} $-\Bigr\{\sum_{h'}
\frac{ \sigma^2(\hat{\Sigma}_{{\bm B}m}^{-1})^{(t+1)}_{h' h} }{k^{(t+1)} Z_B^{(t+1)} }
{\rm erf}\bigr(\omega^{(t+1)}_{h' m}\bigr)
\Bigr\}^2, \ \forall h$}
\STATE{ $\bar{b}_{hm}^{(t+1)} =
\sum_{h',l} (\hat{\Sigma}_{{\bm B} m}^{-1})_{hh'}^{(t+1)} v_{lm} \bar{a}_{lh'}^{(t)}
 - \sum_{h'}
\frac{ \sigma^2(\hat{\Sigma}_{{\bm B}m}^{-1})_{h' h}^{(t+1)}  }{k^{(t+1)} Z_B^{(t+1)}}
{\rm erf}\bigr(\omega_{h' m}^{(t+1)} \bigr),\ \forall h,m$}

\STATE{$t \longrightarrow t+1$}
\ENDWHILE
\STATE{The result of MF is obtained from $\bar{a}_{lh}^{(t)}, \bar{b}_{hm}^{(t)}\ \forall l,h,m$.}
\end{algorithmic}
\end{algorithm}

%%%%%%%%%%%%%%%%%%%%%
\section{Numerical experiment}
\label{sec4}
\begin{figure}
\begin{picture}(0,250)
\put(80,0){
\includegraphics[width=0.6\textwidth,angle=0]{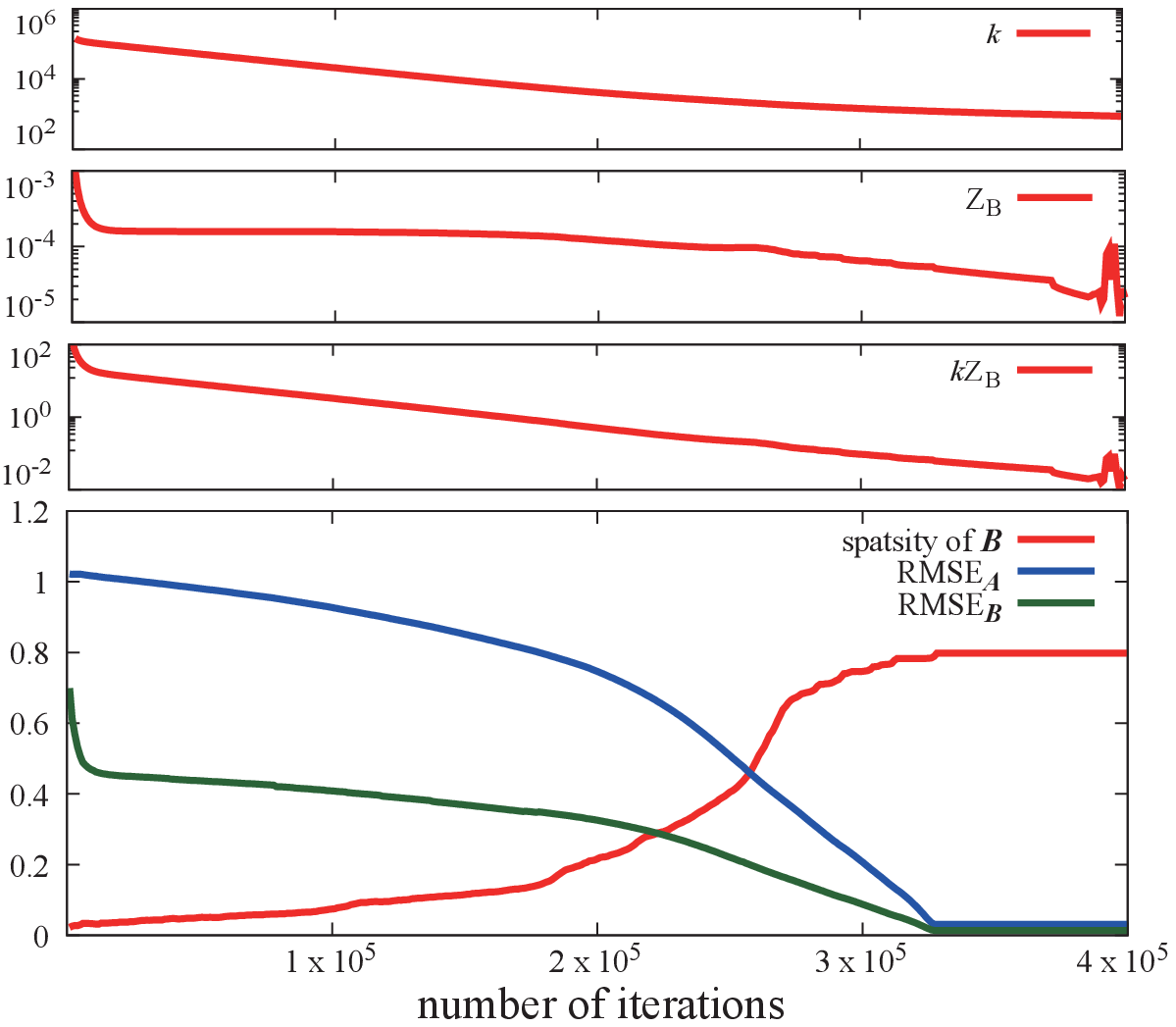}
}
\end{picture}
\caption{A typical dynamical behavior of the proposed sparse MF algorithm.}
\label{fig1}
\begin{picture}(0,340)
\put(70,0){
\includegraphics[width=0.6\textwidth,angle=0]{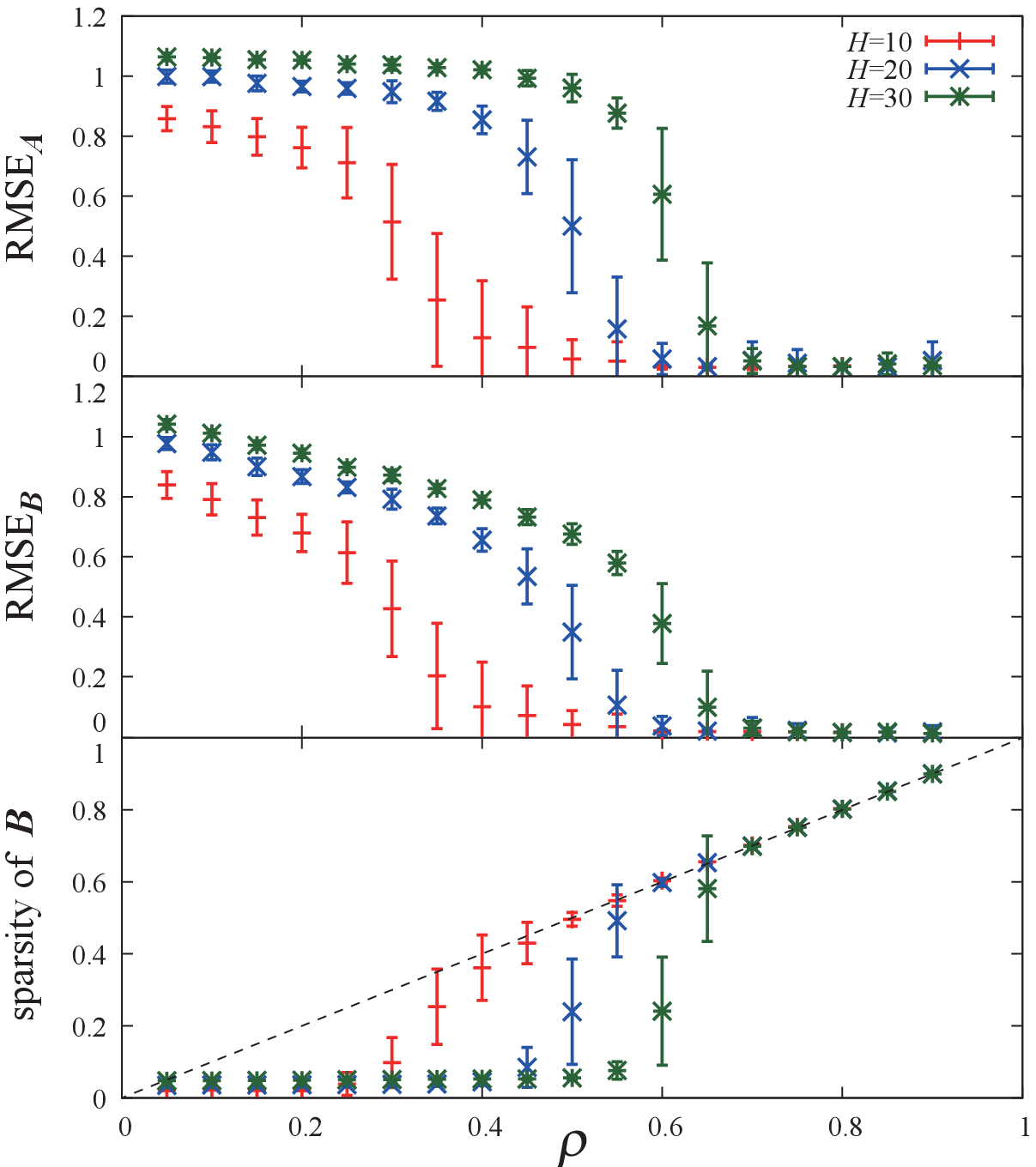}
}
\end{picture}
\caption{The result of sparse matrix reconstruction: 
${\rm RMSE}_{\bm A}$ (top), ${\rm RMSE}_{\bm B}$ (middle), and sparsity of reconstructed $\bm B$ (bottom) are shown.}
\label{fig2}
\end{figure}

\begin{figure}
\begin{picture}(0,150)
\put(80,0){
\includegraphics[width=0.60\textwidth,angle=0]{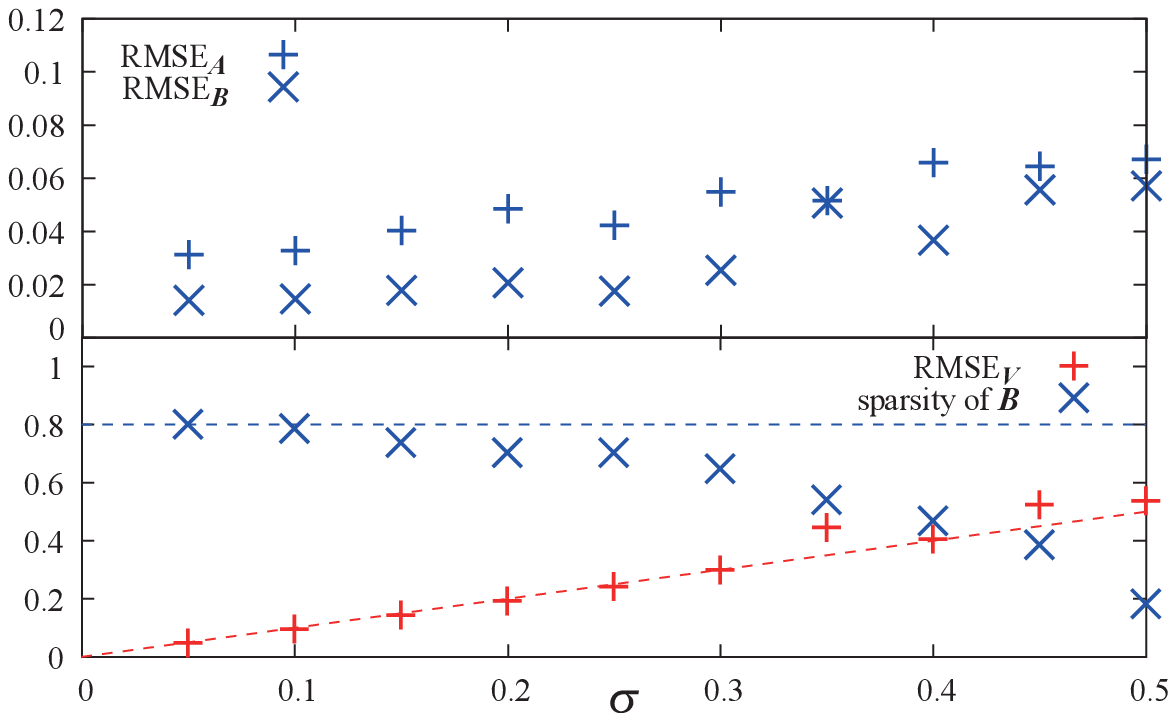}
}
\end{picture}
\caption{The dependence on noise magnitude $\sigma$.}
\label{fig3}
\begin{picture}(0,430)
\put(80,0){
\includegraphics[width=0.60\textwidth,angle=0]{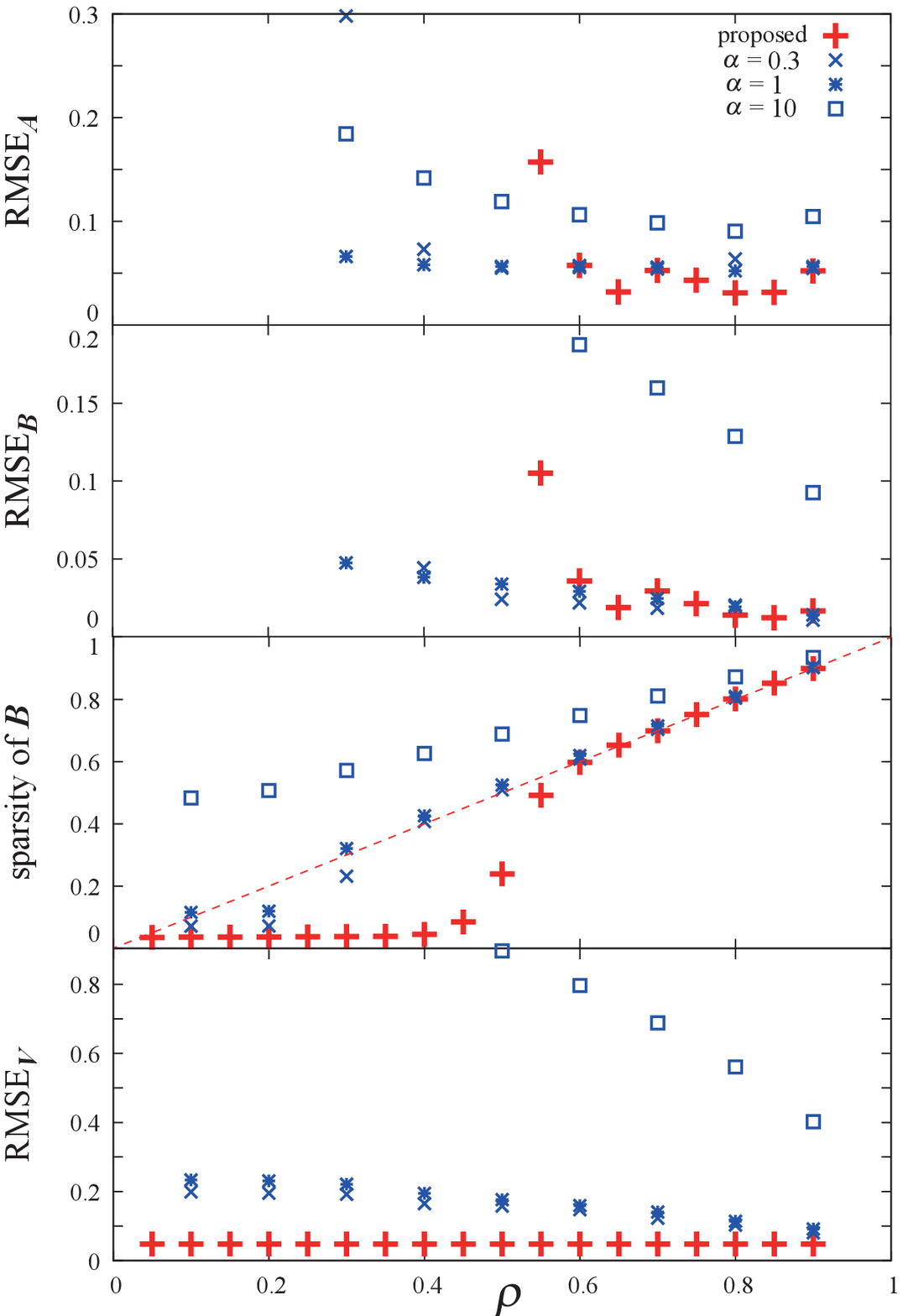}
}
\end{picture}
\caption{Comparison of performance with sparse PCA algorithm in \cite{MBPS}: 
From top to bottom,  
${\rm RMSE}_{\bm A}$, ${\rm RMSE}_{\bm B}$, sparsity of reconstructed $\bm B$, and ${\rm RMSE}_{\bm V}$
are shown.}
\label{fig4}
\end{figure}

For evaluation of sparse MF performance, we numerically reconstruct ground-truth factorized matrix
$\bm A^*, \bm B^*$ by algorithm \ref{alg1}. 
The setup of our numerical experiment for synthetic data is as follows. 
Each element in $\bm A^*, \bm B^*$ is drawn from standard Gaussian distribution
${\cal N} (0, 1)$ and Bernoulli-Gaussian distribution
$P(\bm B^*_{hm}) = (1- \rho) \delta( \bm B^*_{hm} ) + \rho \exp(-(\bm B^*_{hm})^2 / 2) / \sqrt{2\pi}$ $\forall h,m$, respectively. The parameter $\rho$ describes sparsity of $\bm B^*$. The covariance matrix $\bm C_{\bm A}$ is set to be identity.
In the experiments of synthetic data we set $\epsilon=0.1$ and $L=M=500$.
We set $Z_{B{\rm thres}}=10^{-5}$ and the result is averaged over 20 trials excepting figure \ref{fig1}.
Noise magnitude $\sigma$ is set to be $0.05$ in figures \ref{fig1}, \ref{fig2}, and \ref{fig4}.

We should note that sparse MF has degenerate solutions: $\bm V$ is invariant
under permutation of $h$ in $\bm A^*_{lh}, \bm B^*_{hm}$ and
scaling $\{ \bm A^*_{lh}, \bm B^*_{hm}
\} \!\! \rightarrow \!\! \{ c_{h} \bm A^*_{lh}, \bm B^*_{hm} / c_{h} \}$ $ \forall l,m$, where $c_{h}$ is 
 an arbitrary constant.
Therefore, for appropriate measure of ground-truth matrix reconstruction,
we define rooted mean squared error (RMSE) between ground-truth matrices and reconstructed ones as follows.
\begin{eqnarray}
{\rm RMSE}_{\bm A} \!\!\!\!&=&\!\!\!\! \sqrt{ \frac{1}{LH} \sum_{h} \Big( \underset{h',s_h}{\rm min} \sum_{l} \Big( \bm A^*_{lh} 
 - \frac{ s_h} { N_{h'}}  \bar{a}_{l h'} \Big)^2 \Big) }, \\ 
{\rm RMSE}_{\bm B} \!\!\!\!&=&\!\!\!\! \sqrt{ \frac{1}{HM} \sum_{h} \Big( \underset{h'}{\rm min} 
\sum_{m} \Big( \bm B^*_{hm} - s'_h  N_{h'} \bar{b}_{h'm} \Big)^2 \Big) }, 
\end{eqnarray}
where
\begin{eqnarray}
N_{h'} &=& \sqrt{ \frac{ \sum_{l'} (\bar{a}_{l' h'} )^2}{L} }, \\
s'_h &=& \underset{s_h}{\rm argmin}  \Big( \underset{h'}{\rm min} \sum_{l} \Big( \bm A^*_{lh} - 
\frac{ s_h }{ N_{h'}} \bar{a}_{l h'} \Big)^2 \Big).
\end{eqnarray}
The normalization factor $N_{h'}$ is necessary for comparison between reconstructed and ground-truth matrix elements.
Remember that $\ell_2$-norm of column vector in $\bm A^*$ is approximately $\sqrt{L}$ for $L \rightarrow \infty$.
The factor $s_h \in \{ \pm 1 \} $ is for removal of sign ambiguity from scale invariance. 
Minimization with respect to $h'$ is for correct one-to-one correspondence between vectors in ground-truth/reconstructed matrices, because MF has permutation invariance as stated above.
Similarly, for sparsity measure, element in reconstructed $\bm B$ is regarded as sparse if its absolute value is smaller than $10^{-2}$ after scaling the matrix as $\{ \bm \bar{a}_{lh}, \bar{b}_{hm}
\} \!\! \rightarrow \!\! \{ \bar{a}_{lh}/N_h', N_h' \bar{b}_{hm} \}$ $ \forall l,m$ for normalizing $\bm A$.
 Then the sparsity of $\bm B$ is measured by the fraction of sparse element in $\bm B$.

In addition, we also define rooted mean squared error for multiplied matrix
for validity of sparse MF solution,
\begin{equation}
{\rm RMSE}_{\bm V} =  \sqrt{ \frac{1}{LM} \sum_{l,m} \Big( \bm V_{lm} - \sum_{h} \bar{a}_{lh} \bar{b}_{hm} \Big)^2 }.
\end{equation}
${\rm RMSE}_{\bm V}$ will be equal to noise magnitude $\sigma$ when MF is successful.

%%%%%%%%%%%%%%%
First, a typical dynamical behavior of our algorithm under $\rho = 0.8, H = 20$ is depicted in figure \ref{fig1}.
As $Z_B$ gets close to zero, ${\rm RMSE}_{\bm A}$ and ${\rm RMSE}_{\bm B}$ decrease,
and sparsity of $\bm B$ approaches the original value of $\rho$ ($=0.8$). 
This means good reconstruction performance of our algorithm.
At late stage of iteration, unstable behavior of $Z_B$ is observed, therefore 
our algorithm should be terminated before such instability occurs.  
For comparison, we also conduct the experiment under fixed hyperparameter $k$ like in the prior work \cite{KT}, 
where the update of $k$ in algorithm \ref{alg1} is removed.
However, we find that reconstruction performance is poor. 
The MF experiment under fixed $k$ indicates that the sparsity of $\bm B$ in MF solution remains almost constant 
against the change of $k$ even after one million of iterations, 
whose value is much smaller than the ground-truth sparsity $\rho$.
This implies that update of $k$ is essential for ground-truth matrix reconstruction with high accuracy.

Next, we conduct an experiment with sparsity $\rho$ and dimension $H$ being varied. The result in figure \ref{fig2}
indicates the threshold of $\rho$ for crossover to nearly-perfect reconstructable region.  
For larger $H$, many sparse MF solutions other than the ground-truth will exist due to 
larger number of factorized matrix elements, 
which will make ground-truth matrix reconstruction more difficult and lead to larger threshold value.
Note that transition between perfect/imperfect reconstruction phases under noiseless case 
and behavior of RMSE are analyzed in prior works \cite{SK, KMZ, KKMSZ, SSZ, LKZ2, LKZ1}, however not under $\ell_1$ regularizer. 
For robustness to noise, we also observe $\sigma$ dependence under $\rho=0.8, H=20$. In this case, 
standard deviation of element in matrix $\bm A^* \bm B^*$ is evaluated as $2.01 \pm 0.04$
for comparison with $\sigma$ or for signal-to-noise ratio.
The result in figure \ref{fig3} means that ground-truth solution can be found with high accuracy
even in relatively noisy case.

We also compare the performance of our algorithm with widely-used sparse PCA algorithm \cite{MBPS}, where
Lasso for sparse $\bm B$ and dictionary estimation for $\bm A$
 are performed alternately. This is 
implemented in scikit-learn library in Python and easily accessible. In this experiment Python version 2.7 is used.
For sparse PCA, we conduct $10^4$ iterations and vary the coefficient of $\ell_1$ regularizer, denoted by $\alpha$ in figure \ref{fig4}. Default values are used for other hyperparameters. 
We evaluate ${\rm RMSE}_{\bm A}$, ${\rm RMSE}_{\bm B}$, and ${\rm RMSE}_{\bm V}$ for both algorithms under $H=20$ as shown in figure \ref{fig4}.
The result shows that reconstruction performance of our algorithm is better than sparse PCA for larger $\rho$.
Another advantage of our algorithm is that ${\rm RMSE}_{\bm V}$ is kept constant
and almost equal to $\sigma$ (0.05 in the present case) regardless of the value $\rho$, while ${\rm RMSE}_{\bm V}$ increases for smaller $\rho$ in sparse PCA.
In our algorithm, we assume that the value of noise magnitude $\sigma$ is known in advance, 
which will lead to constant ${\rm RMSE}_{\bm V}$. However, it should be emphasized that
such constant behavior of ${\rm RMSE}_{\bm V}$ is nontrivial because VB solution for sparse MF 
was obtained under several approximations.

%%%%%%%%%%%%%%%
Finally, for verifying practical utility of our algorithm, 
we also conduct an experiment for extraction of dictionary in real image. In this experiment,
we use four monochrome images of $256 \times 256$ pixels ($L=M=256$) in Volume 3
in The USC-SIPI Image Database \cite{database}, namely
Tree (image \# 4.1.06), Moon Surface (5.1.09), Aerial (5.1.10), and Clock (5.1.12).
Before applying MF, the values of each pixel are normalized to have
zero mean and one variance, which is regarded as the observed matrix $\bm V$
in our formulation.
For comparison, we apply two MF methods, our algorithm and sparse PCA algorithm.
In applying our algorithm to real image we set $\epsilon=10^{-3}$, $Z_{B{\rm thres}}=10^{-5}$ and the result is averaged over 5 trials
by changing initial random factorized matrices. For sparse PCA, we conduct $10^4$ iterations and vary the $\ell_1$ regularizer coefficient $\alpha$. The results of ${\rm RMSE}_{\bm V}$ and sparsity of $\bm B$ are evaluated for both algorithms.
 
\begin{figure}
\begin{picture}(0,550)
\put(0,0){
\includegraphics[width=1\textwidth,angle=0]{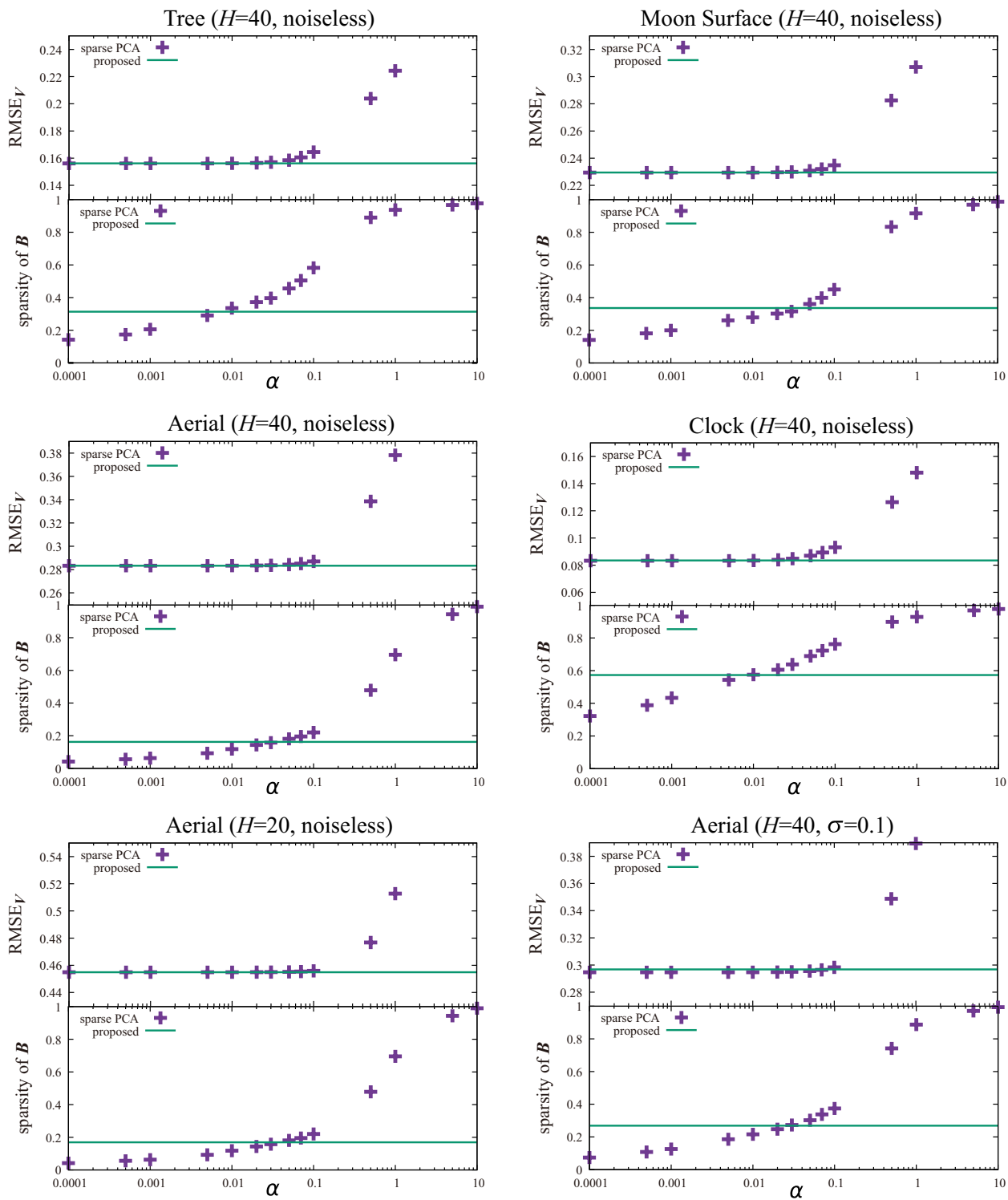}
}
\end{picture}
\caption{Comparison of performance with sparse PCA algorithm for real image: 
The results for Tree (top left, $H=40$, noiseless), Moon Surface (top right, $H=40$, noiseless),
Aerial (middle left, $H=40$, noiseless), Clock (middle right, $H=40$, noiseless),
Aerial under smaller $H$ (bottom left, $H=20$, noiseless), and noisy Aerial (bottom right, $H=40$, $\sigma=0.1$)
are shown.}
\label{fig5}
\end{figure}

The result is shown in figure \ref{fig5}.
We mainly conduct the experiment under $H=40$ and without noise. In the application to the noiseless case, we cannot set the noise parameter $\sigma=0$ in our algorithm, then we set $\sigma=0.03$ for the MF solution. In sparse PCA, the results for all four images indicate that ${\rm RMSE}$ decreases for smaller $\alpha$ and has almost a constant minimum value 
below a certain value of $\alpha$, whereas the sparsity of $\bm B$ constantly decreases. 
In contrast, almost minimum ${\rm RMSE}_{\bm V}$ is obtained by our algorithm without tuning hyperparameter. 
Furthermore, it appears that the sparsity of $\bm B$ by our algorithm is close to 
the largest value within the region of almost-constant minimum ${\rm RMSE}_{\bm V}$ by sparse PCA.
Such behaviors do not change even under different $H$ ($H=20$) or the noisy case ($\sigma=0.1$ in $\bm E$, 
where we also set $\sigma=0.1$ in our algorithm).

This result implies that our algorithm might be able to find the MF solution having the nearly sparsest $\bm B$ 
within the region of nearly minimum ${\rm RMSE}_{\bm V}$.
 This is the significant advantage of our algorithm because we do not 
need to tune hyperparameter for sparse MF, namely the coefficient of $\ell_1$ regularizer.
In addition, this result also suggests that our algorithm excellently works even for real data. This fact is nontrivial because 
the elements in factorized matrices $\bm A, \bm B$ should be correlated in real image, 
whereas we assume i.i.d prior for $\bm A, \bm B$ in our formulation.

%%%%%%%%%%%%%%%%%%%%%
\section{Summary}
\label{sec5}

We proposed a sparse MF algorithm including hyperparameter tuning in Laplace prior.
Surprisingly, although VB solution for sparse MF is derived under several approximations, 
our algorithm is successful for finding ground-truth sparse MF solution with high accuracy.
We also found that our algorithm shows the excellent performance for extracting dictionary 
in real image.

Here we only numerically verified the performance of our algorithm, and
several problems remain unresolved.
For example, we need to analyze reconstruction performance and convergence condition theoretically. However it will be difficult
due to the complex expression of VB solution. The analysis of VB solution under Gaussian prior \cite{IR,NS},
which is much simpler than the present case, will help us to understand performance and dynamics of our algorithm.
Next problem is partial update of hyperparameter for convergence.
In general, very small update parameter $\epsilon$ leads to stable convergence, while it 
makes the algorithm very slow.
Therefore, we need to find the strategy for faster convergence.
Finally, our method for hyperparameter tuning is completely novel, and
future application may give new insights to other sparse modeling problems.

%%%%%%%%%%
\section*{Acknowledgments}

We appreciate comments from Tomoki Tamai.
This work is supported by KAKENHI Nos. 18K11175, 19K12178, 20H05774, and 20H05776.
%%%%%%%%%%

%\begin{thebibliography}{99}%
\bibliographystyle{unsrt}
\bibliography{article}

%\end{thebibliography}%

\end{document}